\documentclass[sigconf,natbib=false,nonacm]{acmart}
\AtBeginDocument{%
  }

\setcopyright{acmlicensed}

\copyrightyear{2026}
\acmYear{2026}
\acmDOI{XXXXXXX.XXXXXXX}
\acmConference[SAC'26]{The 41st ACM/SIGAPP Symposium on Applied Computing}{March 23--27, 2026}{Thessaloniki, Greece}
\acmISBN{979-X-XXXX-XXXX-X/26/03}

\acmSubmissionID{3124}

\RequirePackage[
  datamodel=acmdatamodel,
  style=acmnumeric,
  ]{biblatex}

\usepackage{subcaption}
\usepackage{comment}
\usepackage{booktabs}
\usepackage{color, colortbl}
\usepackage{float}
\usepackage{multirow}
\usepackage{xcolor}

\definecolor{customorange}{RGB}{255,228,202}
\definecolor{customblue}{RGB}{224,234,246}

\begin{document}

\title{Learning to Navigate Under Imperfect Perception:\\ Conformalised Segmentation for Safe Reinforcement Learning}


\renewcommand{\shorttitle}{Learning to Navigate Risky Terrain}
\newcommand{\approach}{COPPOL}
\author{Daniel Bethell}
\orcid{0000-0002-0685-5312}
\affiliation{
  \institution{University of York}
  \city{York}
  \country{UK}
}
\email{daniel.bethell@york.ac.uk}

\author{Simos Gerasimou}
\affiliation{%
  \institution{University of York}
  \city{York}
  \country{UK}
}
\email{simos.gerasimou@york.ac.uk}

\author{Radu Calinescu}
\affiliation{%
  \institution{University of York}
  \city{York}
  \country{UK}
}
\email{radu.calinescu@york.ac.uk}

\author{Calum Imrie}
\affiliation{%
  \institution{University of York}
  \city{York}
  \country{UK}
}
\email{calum.imrie@york.ac.uk}

\renewcommand{\shortauthors}{D. Bethell et al.}

\begin{abstract}
Reliable navigation in safety-critical environments requires both accurate hazard perception and principled uncertainty handling to strengthen downstream safety handling. 
Despite the effectiveness of existing approaches, they assume perfect hazard detection capabilities, while uncertainty-aware perception approaches lack finite-sample guarantees. 
We present \approach, a conformal-driven perception-to-policy learning approach that integrates \emph{distribution-free, finite-sample safety guarantees} into semantic segmentation, yielding calibrated hazard maps with rigorous bounds for missed detections.
These maps induce risk-aware cost fields for downstream RL planning. Across two satellite-derived benchmarks, \approach\ increases hazard coverage (up to $6 \times$) compared to comparative baselines, achieving near-complete detection of unsafe regions while reducing hazardous violations during navigation (up to $\approx 50\%$). 
More importantly, our approach remains robust to distributional shift, preserving both safety and efficiency.
\end{abstract}



\maketitle


\section{Introduction}
\label{sec:Introduction}

Deploying robots in remote, safety-critical, and resource-intensive environments requires navigation capabilities that surpass human limitations~\cite{guiochet2017safety}, a role increasingly fulfilled by autonomous systems~\cite{chen2022unmanned}. 
Unlike humans, these systems are not subject to fatigue, distraction, or limited perceptual bandwidth~\cite{riener2023robots}, and can operate continuously in conditions that would rapidly overwhelm human operators, such as planetary exploration~\cite{gao2016contemporary}, disaster response~\cite{bahadori2005autonomous}, or subsea inspection~\cite{mai2016subsea}. 
By leveraging advanced sensing and decision-making capabilities, autonomous agents can process  streams of visual and spatial data in real time, optimising their behaviour to achieve mission objectives with greater consistency and safety~\cite{zhi2019learning}. However, the autonomy that enables operation in such challenging settings also imposes stringent requirements. 
When perception is noisy, incomplete, or uncertain, even minor errors in hazard detection can cascade through decision-making, resulting in catastrophic failures~\cite{qutub2022hardware}. Ensuring that autonomous navigation is not only effective but also provably safe remains a central challenge for real-world deployment.

Recent research introduces approaches that partially address these challenges through RL extensions that incorporate explicit safety constraints. Safe RL methods, including constrained optimisation~\cite{achiam2017constrained, zhaostate}, Lyapunov-based techniques~\cite{chow2018lyapunov, berkenkamp2017safe}, and shielding~\cite{alshiekh2018safe, bethell2024safe, carr2023safe}, can prevent unsafe actions during training and execution. 
However, these approaches assume that hazards are perfectly captured in the agent’s observations, an assumption that rarely holds in practice. 
To mitigate the risks of perceptual error, uncertainty-aware~\cite{abdar2021review} navigation has been proposed, where Bayesian models~\cite{goan2020bayesian} or Monte Carlo dropout~\cite{gal2016dropout, bethell2024robust} quantify predictive uncertainty and propagate it into mapping and planning~\cite{banfi2022worth}. 
While such methods reduce overconfidence, their probabilistic estimates remain heuristic and lack finite-sample safety guarantees. 
Conformal prediction (CP)~\cite{shafer2008tutorial, angelopoulos2023conformal} offers a principled alternative, yielding distribution-free coverage guarantees for perception tasks such as segmentation~\cite{mossina2024conformal} and detection~\cite{timans2024adaptive}. 
Yet, existing work has treats perception and decision-making separately: CP-based segmentation ensures hazards coverage without informing the policy, while CP-based decision theory calibrates actions without addressing upstream perceptual errors. 

We mitigate this gap by introducing \approach, a unified conformal-driven perception-to-policy learning approach for safe navigation that integrates conformal prediction~\cite{angelopoulos2022conformal} directly into the perception stream to inform downstream policy learning.
Given an overhead view of the terrain (e.g., satellite imagery around an autonomous rover), a segmentation Deep Neural Network (DNN) produces per-pixel hazard scores, which we calibrate into set-valued hazard maps with finite-sample guarantees on missed hazards. The resulting calibrated mask is projected on the scene to induce a risk-aware cost field, over which an RL agent plans its trajectory. 
By reasoning over statistically guaranteed hazard coverage instead of uncalibrated probabilities, the RL agent learns policies that are both effective  and statistically safe. 
Through our extensive evaluation on hazardous-terrain benchmarks, we demonstrate that \approach\ consistently improves both perception and navigation safety. 
In particular, our approach increases hazard coverage by more than 6x compared to state-of-the-art baselines, achieving highly-accurate detection ($\geq 90\%$) of unsafe regions on some benchmarks.
Furthermore, RL agents spend approximately 50\% time less in hazardous areas and maintain two to three times greater distance (clearance) from unsafe terrain. 
These results show that conformal segmentation enables substantially safer trajectories without sacrificing task efficiency. 

Our key contributions are as follows: (1) a conformal segmentation process that converts raw imagery into set-valued hazard maps with finite-sample guarantees on coverage; (2) \approach, a perception-to-policy approach that leverages these calibrated maps to enable RL agents to reason over a risk-aware representation of their environment; and (3) an extensive empirical evidence showing that our approach increases hazard coverage by more than sixfold, reduces unsafe incidents by approximately 50\%, and maintains two to three times greater clearance (distance) from hazards compared to baselines, yielding substantially safer navigation without compromising task efficiency.


\section{Related Work}
\label{sec:Related Work}

\textbf{Safe Reinforcement Learning.} 
A core line of research in safe autonomous navigation extends the standard RL framework to explicitly incorporate safety constraints~\cite{wachi2024survey}.
Constrained Policy Optimisation introduced guarantees on constraint satisfaction by employing a trust-region update that maximises return while enforcing cost limits through Lagrange multipliers~\cite{achiam2017constrained, zhaostate}.
Lyapunov-based approaches extend this idea by constructing Lyapunov functions that bound long-term constraint violations, yielding policies that remain feasible throughout training~\cite{berkenkamp2017safe,chow2018lyapunov}.
Beyond optimisation-based methods, safety layers~\cite{dalal2018safe} and safety shields~\cite{alshiekh2018safe, bethell2024safe, carr2023safe} intervene during execution to prevent the policy from taking unsafe actions, suggesting alternative and feasible actions.  
Despite methodological differences, these approaches assume that the agent’s observations faithfully capture all relevant hazards. 
When, however, perception is incomplete, such as when an observation model (e.g., a convolutional neural network)  fails to detect a hazard/object, safety cannot be enforced 
as the policy lacks the information required to constrain action selection against unseen risks.
Thus, safe RL alone is insufficient for environments with  high perceptual uncertainty.


\textbf{Uncertainty-Aware Navigation.}
Uncertainty Quantification (UQ)~\cite{abdar2021review} aims to quantify and propagate perceptual uncertainty through the autonomous navigation stack.
Bayesian deep networks \cite{goan2020bayesian} and their scalable approximations, like Monte Carlo dropout~\cite{gal2016dropout, bethell2024robust} have been applied to perception and planning systems~\cite{kendall2015bayesian, flogel2024disentangling} to represent epistemic ambiguity.
 These uncertainty estimates can also be injected into mapping frameworks, for instance, by augmenting occupancy grids with per-cell confidence scores, enabling planners to balance path optimality against perceptual reliability~\cite{banfi2022worth}. 
These uncertainty-aware approaches, albeit avoiding overconfident navigation decisions, remain inherently probabilistic: 
calibrated probabilities or variance estimates serve as useful heuristics but do not provide finite-sample guarantees.
In practice, this implies that even when a model declares high uncertainty, there is no assurance that critical hazards are captured within the uncertainty bounds, leaving residual risk of catastrophic oversight.


\textbf{Conformal Prediction for Navigation.} 
Conformal prediction (CP)~\cite{shafer2008tutorial, angelopoulos2023conformal} is a distribution-free statistical framework for quantifying predictive uncertainty with finite-sample coverage guarantees.
In computer vision, CP has been applied to semantic segmentation by constructing pixel-wise prediction sets that provably bound the false-negative rate of critical classes~\cite{mossina2024conformal, angelopoulos2022conformal}, ensuring that hazardous regions are included with high probability. 
In object detection, these guarantees are extended to bounding boxes, where conformal detectors calibrate outputs to control the likelihood of missed detections~\cite{andeol2023confident, timans2024adaptive}. 
Beyond perception, conformal decision theory~\cite{lekeufack2024conformal} has been proposed to calibrate the decision-making process itself, yielding policies that maintain desired risk levels despite imperfect predictions.


However, existing work has largely treated perception and decision-making in isolation.
Conformal decision theory calibrates action selection 
to satisfy global risk bounds but leaves the perceptual predictions unchanged, yielding coarse decision-level guarantees and lacking spatial awareness of hazards.
Consequently, approaches within this framework often default to conservative fallback strategies that degrade efficiency without ensuring that critical risks are captured upstream. 
This persistent disconnect between perception and control motivates a unified treatment of uncertainty across the navigation stack.
Thus, in \approach, conformal prediction is integrated directly into hazard perception, producing pixel-level coverage guarantees that are propagated to the RL agent.
By coupling conformally calibrated segmentation with safe control, our method bridges the gap between perception and decision-making, enabling navigation policies that are both risk-aware and grounded in statistically valid safety guarantees.



\begin{figure*}[tb]
    \centering
    \includegraphics[width=0.95\linewidth]{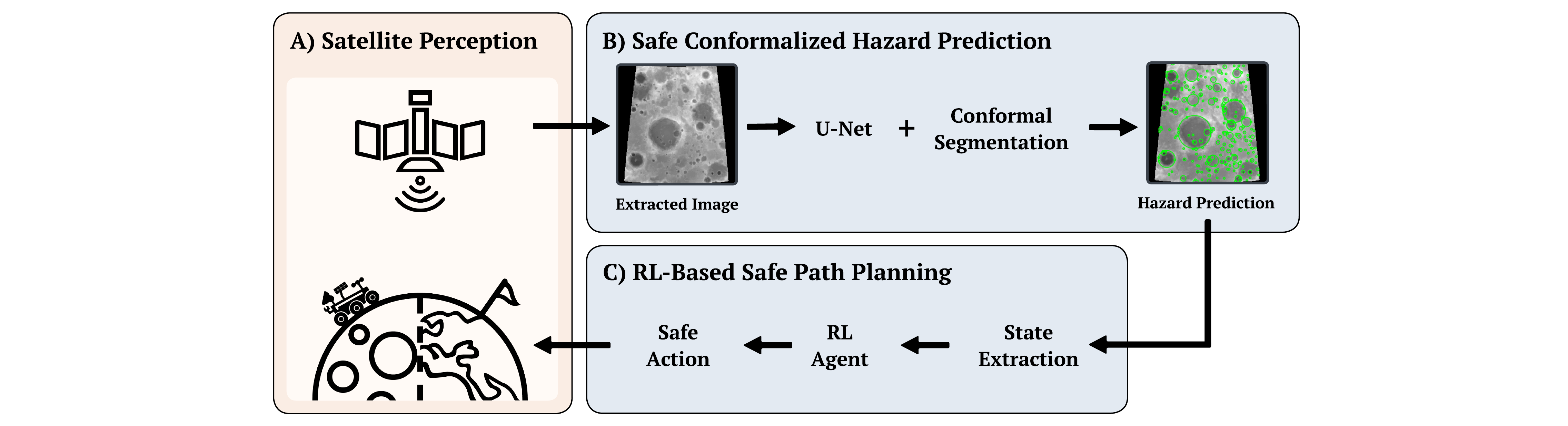}
    \vspace{-3mm}
    \caption{High-level overview of \approach, our conformal-driven perception-to-policy learning approach for safe navigation over risky terrain. \approach\ combines 
    \textbf{(A)} real-world image capture via satellites of the rover's surrounding terrain, \textbf{(B)} U-Net-based segmentation of hazards with uncertainty calibration through conformal prediction, and  \textbf{(C)} an RL agent to plan a safe trajectory to the goal. \textcolor{orange}{Orange} modules represent the environment; \textcolor{blue}{blue} modules are learned components of our approach.}
    \label{fig:framework-overview}
    \Description[A high-level overview of our proposed framework.]{A high-level overview of our proposed framework.
    }
    \vspace{-3mm}
\end{figure*}

\section{Preliminaries}
\label{sec:Preliminaries}

\textbf{Markov Decision Process and Reinforcement Learning.} A Markov Decision Process (MDP)~\cite{bellman1957markovian} is a discrete-time stochastic control process for modelling decision-making tasks. An MDP is formally defined as a 5-tuple $M = (S, A, P, R, \gamma)$, where $S$ is the state space, $s_t$ is the state at timestep $t$, $A$ is the action space, $P$ is the state transition probability matrix/transition function such that $P(s_{t+1}|s_t, a_t)$ is the probability of transitioning to state $s_{t+1}$ from state $s_t$ using action $a_t$, $R$ is the reward function such that $R(s,a)$ is the reward for taking action $a$ in state $s$, and $\gamma$ is the discount factor that determines the weight of future rewards. A policy $\pi:  S \to\Delta(A)$ is a distribution over actions given a state. 

Reinforcement Learning (RL) involves training an agent to make a sequence of decisions by interacting with the environment, which is modelled as an MDP. This machine learning paradigm solves MDPs when full knowledge of the transition function is not available. The agent's goal is to find a policy $\pi^*$ maximising the expected discounted return $E\left[\sum_{t=0}^{\infty}\gamma^{t}R(s_{t}, a_{t})\right]$~\cite{sutton2018reinforcement}. A value function $V_{\pi}(s)$ informs the agent how \textit{valuable} a given state is when following the current policy $\pi$.

\textbf{Conformal Prediction.} Deep neural networks produce point or probabilistic estimates that may be miscalibrated and fail to capture epistemic or aleatoric uncertainty~\cite{abdar2021review}. Standard approaches to uncertainty quantification (e.g., Bayesian neural networks~\cite{goan2020bayesian} or Monte Carlo dropout~\cite{gal2016dropout}) provide useful heuristics but provided reduced coverage guarantees. 
Conformal prediction (CP), instead, is a distribution-free framework for uncertainty quantification underpinned by finite-sample coverage guarantees. 
Given a nonconformity score function $\ell: \mathcal{X} \times \mathcal{Y} \to \mathbb{R}$, a calibration dataset $\mathcal{D}_{\text{cal}} = \{(x_i, y_i)\}_{i=1}^n$, and a risk parameter $\alpha \in (0,1)$, CP constructs a prediction set (for classification)~\cite{angelopoulos2020uncertainty} or prediction interval (for regression)~\cite{romano2019conformalized} $\mathcal{C}(x_{n+1})$ for input $x_{n+1}$ that satisfies:
\begin{equation}
P\bigl( y_{n+1} \in \mathcal{C}(x_{n+1}) \bigr) \geq 1 - \alpha
\end{equation}
under the standard exchangeability assumption between calibration and test data~\cite{angelopoulos2023conformal}.


\section{Approach}
\label{sec:Methodology}

\approach\ is a conformal-driven perception-to-policy learning approach (Figure~\ref{fig:framework-overview}) that converts raw imagery into calibrated hazard maps for downstream navigation. 
\approach\ comprises (A): real-world imagery captured via satellites of a rover's local terrain, (B): a convolutional U-Net that segments hazardous terrain at the pixel level with uncertainty calibration through conformal prediction, and (C): an RL agent that plans safe trajectories across the risky terrain.\footnote{We use space navigation as a motivating example, but \approach\ can comfortably generalise to any perception-to-policy pipeline.} 
Unlike traditional pipelines where the RL agent consumes raw or thresholded segmentation outputs, \approach\ treats calibrated hazard regions as integral information, allowing the policy to reason over safety-aware spatial representations rather than uncalibrated predictions. This decoupling of perception and planning under explicit risk guarantees enables principled navigation in environments where perception errors can have harmful consequences.

\subsection{Conformalized Hazard Prediction}
\label{sec:Conformalized Hazard Prediction}
We formalise hazard perception as set-valued, distribution-free inference over the spatial domain of an observation. Each observation is an image $x \in \mathbb{R}^{H \times W \times C}$, accompanied by a binary hazard mask $y \in \{0,1\}^{H \times W}$ marking unsafe terrain. 
The image is defined over the pixel lattice $\Omega = \{1, \ldots, H\} \times \{1, \ldots, W\}$, where the true hazardous set for the $n$-th image is $\mathcal{H}_n = \{ (i,j) \in \Omega : y_{i,j}=1 \}$.

Hazard scores are produced by a convolutional U-Net~\cite{ronneberger2015u} $f_\theta: \mathbb{R}^{H \times W \times C}$ $\rightarrow [0,1]^{H \times W}$ trained on the dataset $\mathcal{D}_{\text{train}}$ with a binary cross-entropy loss. 
The U-Net network outputs the per-pixel probability of being hazardous, $p_\theta(x)$. 
This step enables transforming raw imagery into a dense field of hazard probabilities, which is then calibrated into explicit, risk-controlled hazard regions. 

The baseline U-Net employs a single threshold $\lambda \in [0,1]$, chosen a priori (e.g., by validation tuning), and predictions are applied directly to the hazard scores $p_\theta(x)$ to produce the final binary mask.
However, this decision locks the system to an operating point on the precision–recall curve and offers no distribution-free guarantee on missed hazards which is the main quantity of interest. 
\approach\ improves this ad-hoc choice through a principled, risk-controlled mechanism that exposes the entire set of threshold values and considering the nested family of candidate hazard regions:
\begin{equation}
\begin{split}
C_\lambda(x) &= \{ (i,j) \in \Omega : p_\theta(x)_{(i,j)} \geq 1 - \lambda \}, \\
M_{\lambda}(x)_{(i,j)} &= \mathbf{1}\{ (i,j) \in C_\lambda(x) \}
\end{split}
\label{eq:}
\end{equation}
As $\lambda$ increases, $C_\lambda(x)$ expands monotonically, progressively covering more of the terrain. In the context of risky-terrain perception and navigation, the primary objective is to minimise missed hazards. To achieve this under a distribution-free guarantee, we frame the problem within the conformal prediction paradigm, using the pixel-level false negative rate (FNR) as the nonconformity score. 
For an image $(x_n, y_n)$ with hazardous set $\mathcal{H}_n$, we define:
\begin{equation}
L^{\text{FNR}}_n(\lambda) = 1 - \frac{|\mathcal{H}_n \cap C_\lambda(x_n)|}{|\mathcal{H}_n|}
\label{eq:}
\end{equation}
where $L^{\text{FNR}}_n(\lambda) = 0$ if $|\mathcal{H}_n| = 0$. 
This equation measures the fraction of hazardous pixels that the prediction fails to capture.

To determine the operating threshold $\lambda$ in a principled way, we perform calibration on the held-out calibration dataset $\mathcal{D}_{\text{cal}}$.\footnote{Train, calibration, and test sets are strictly disjoint to preserve statistical validity.} 
For any $\lambda \in [0,1]$, the empirical risk with respect to the FNR metric is:

\begin{equation}
\hat{R}_N(\lambda) = \frac{1}{N}\sum^n_{n=1}L^{\text{FNR}}_n(\lambda) \qquad N = |\mathcal{D}_{\text{cal}}|
\label{eq:}
\end{equation}

The conformal procedure searches across the nested family $\{C_\lambda\}$ to find the least conservative threshold $\hat{\lambda}$ that achieves the target risk level $\alpha \in (0,1)$, given by~\cite{angelopoulos2022conformal}:
\begin{equation}
\hat{\lambda} = \inf \biggl\{ \lambda \in  [0,1] : \frac{N}{N+1}\hat{R}_N(\lambda) + \frac{1}{N+1} \leq \alpha \biggr\}
\label{eq:}
\end{equation}

Once $\hat{\lambda}$ is obtained, it is fixed and applied uniformly at test time to transform the probability map $p_\theta(x)$ into a binary, calibrated hazard mask $M_{\hat{\lambda}}(x)$. Under the standard exchangeability assumption between calibration and test samples, this procedure guarantees that the expected pixel-level FNR on unseen data satisfies:
\begin{equation}
\mathbb{E}\bigg[ L^{\text{FNR}}_{N+1}(\hat{\lambda}) \bigg] \leq \alpha
\label{eq:}
\end{equation}

\subsection{Safe RL-Based Path Planning}
\label{sec:Safe Path Planning}
Our navigation problem is posed on the captured image itself. Given an image $x \in \mathbb{R}^{H \times W \times C}$ and its calibrated hazard mask $M_{\hat{\lambda}}(x)$, we construct a planning image by overlaying $M_{\hat{\lambda}}(x)$ on $x$. 
This image is the world the rover must traverse: pixels depict traversable terrain, while mask-marked regions encode terrain that is \emph{predicted risky with distribution-free control of missed hazards}. 
Examples of this planning image can be seen in Figure~\ref{fig:qualitative_comparison}. 
Accordingly, conformal perception does not merely decorate the image; it \emph{induces a distribution-free, risk-calibrated cost field} on the image, which the agent optimises against.

We formalise the task as an MDP on the pixel lattice $\Omega$. The RL agent's pose at time $t$ is the lattice coordinate $q_t = (i_t, j_t) \in \Omega$ with $q_0 \in \Omega$ being the initial coordinate and goal $g \in \Omega$. The action space is then the cardinal directions from the current pixel:
\begin{equation}
A=\{\uparrow,\rightarrow,\downarrow,\leftarrow\},\quad
\begin{aligned}
&\Delta_\uparrow=(-1,0), \quad \Delta_\rightarrow=(0,1),\\
&\Delta_\downarrow=(1,0), \quad \Delta_\leftarrow=(0,-1)
\end{aligned}
\label{eq:}
\end{equation}

and a run concludes 
when $q_{t+1} = g$ or $t=T$.

The MDP state $s_t$ presented to the agent can be either the planning image (e.g., the satellite image with the conformal hazard mask overlaid) or a compact feature vector encoding the same calibrated geometry: 
the goal offset $(D_{i_t}, D_{j_t})$ given by the Manhattan distance to the goal, 
the distance to the nearest predicted hazard rim, 
a binary “inside-hazard” indicator, a
nd four action-legality flags (up/right/down/left) signifying whether an action is allowed or not. 
The ground truth state $q_t$ is only used to compute these quantities; the agent never receives $q_t$ directly. 
In both representations, the agent perceives the same conformally calibrated world, either as pixels or as structured features. 
The choice depends on prioritising between spatial reasoning from raw visual inputs (image view) or learning efficiency and reduced sample complexity (vector view).

\approach\ guides the agent with the Manhattan heuristic $h_t := ||q_t - g||_1$, yielding higher rewards when the agent approaches the goal $g$ while imposing calibrated costs on predicted hazards:
\begin{equation} 
r_t =
\begin{cases}
-1 - \kappa, & \text{if } q_{t+1} \in M_{\hat{\lambda}}(x), \\
\operatorname{sgn}\!\big(h_t - h_{t+1}\big) - \kappa + \beta \,\mathbf{1}\{ q_{t+1} = g \}, & \text{otherwise},
\end{cases}
\label{eq:} 
\end{equation}

with small step cost $\kappa = 0.05$ and a goal reaching bonus $\beta = 50$; $\operatorname{sgn}$ is the signum function, returning 1 for any positive number, -1 for any negative number, and 0 otherwise. 
The first case incurs a conformally-based penalty when entering or standing in regions predicted as risky, and the second case rewards $A^*$-like progress towards the goal while keeping paths short.

The objective of RL planning is to maximise
the discounted return $J(\pi)$ where the agent observes $s_t$ (either the image or vector view derived from $x$, $M_{\hat{\lambda}}(x)$, and $q_t$) and select $a_t \sim \pi(a|s_t)$. 
Our \approach\ approach is agnostic to the discrete-control algorithm: value-based methods that estimate $Q(s,a)$ and act via $\text{argmax}_a Q(s_t,a)$, or a categorical actor that outputs a distribution $\pi(a|s_t)$ over $A$ are equally applicable. 
In our experimental evaluation, we use a convolutional policy for image states and an MLP for vector states.


\section{Evaluation}
\label{sec:Evaluation}

\begin{figure*}[tb!]
\centering
\begin{minipage}{\textwidth}
    \centering
    \begin{subfigure}[t]{0.15\textwidth}
        \includegraphics[width=\textwidth]{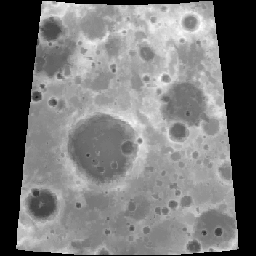}
        \vspace{-5mm}
        \caption*{Input}
    \end{subfigure}
    \hfill
    \begin{subfigure}[t]{0.15\textwidth}
        \includegraphics[width=\textwidth]{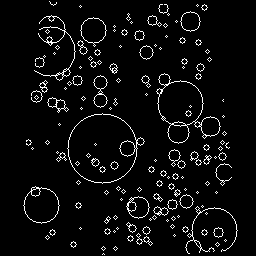}
        \vspace{-5mm}
        \caption*{Ground Truth}
    \end{subfigure}
    \hfill
    \begin{subfigure}[t]{0.15\textwidth}
        \includegraphics[width=\textwidth]{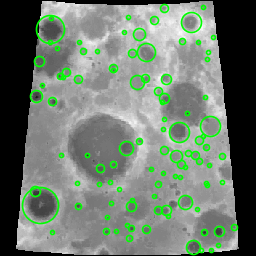}
        \vspace{-5mm}
        \caption*{Baseline}
    \end{subfigure}
    \hfill
    \begin{subfigure}[t]{0.15\textwidth}
        \includegraphics[width=\textwidth]{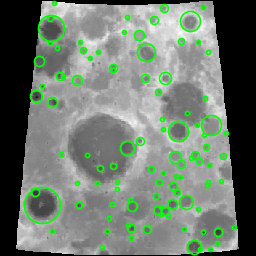}
        \vspace{-5mm}
        \caption*{MC}
    \end{subfigure}
    \hfill
    \begin{subfigure}[t]{0.15\textwidth}
        \includegraphics[width=\textwidth]{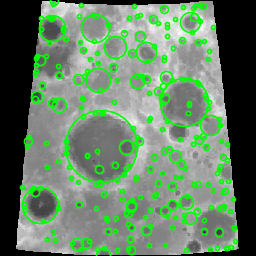}
        \vspace{-5mm}
        \caption*{CRC}
    \end{subfigure}
    \hfill
    \begin{subfigure}[t]{0.15\textwidth}
        \includegraphics[width=\textwidth]{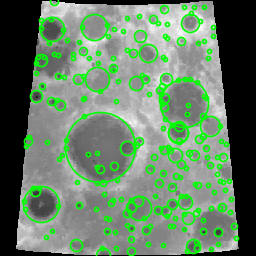}
        \vspace{-5mm}
        \caption*{MC-CP}
    \end{subfigure}
    \vspace{0.75em}
    \makebox[\textwidth]{\textbf{(a) DeepMoon Crater Detection}}
\end{minipage}

\begin{minipage}{\textwidth}
    \centering
    \begin{subfigure}[t]{0.15\textwidth}
        \includegraphics[width=\textwidth]{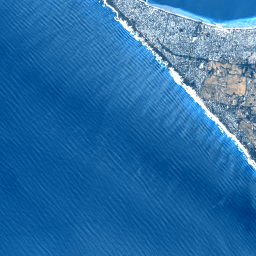}
        \vspace{-5mm}
        \caption*{Input}
    \end{subfigure}
    \hfill
    \begin{subfigure}[t]{0.15\textwidth}
        \includegraphics[width=\textwidth]{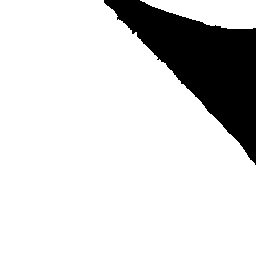}
        \vspace{-5mm}
        \caption*{Ground Truth}
    \end{subfigure}
    \hfill
    \begin{subfigure}[t]{0.15\textwidth}
        \includegraphics[width=\textwidth]{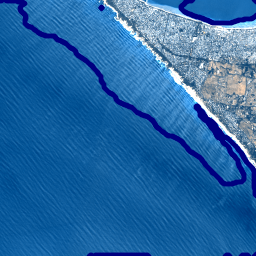}
        \vspace{-5mm}
        \caption*{Baseline}
    \end{subfigure}
    \hfill
    \begin{subfigure}[t]{0.15\textwidth}
        \includegraphics[width=\textwidth]{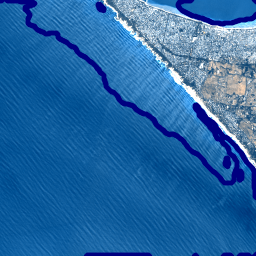}
        \vspace{-5mm}
        \caption*{MC}
    \end{subfigure}
    \hfill
    \begin{subfigure}[t]{0.15\textwidth}
        \includegraphics[width=\textwidth]{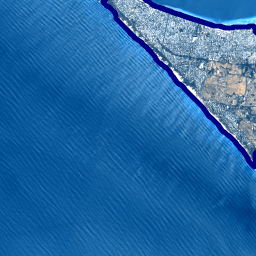}
        \vspace{-5mm}
        \caption*{CRC}
    \end{subfigure}
    \hfill
    \begin{subfigure}[t]{0.15\textwidth}
        \includegraphics[width=\textwidth]{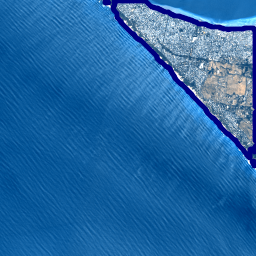}
        \vspace{-5mm}
        \caption*{MC-CP}
    \end{subfigure}
    \vspace{0.75em}
    \makebox[\textwidth]{\textbf{(b) YTU-WaterNet Water Detection}}
\end{minipage}
\vspace{-1.5em}
\caption{Comparison of hazard predictions for (a) DeepMoon and (b) YTU-WaterNet hazards. Each row shows the input image captured via satellite, ground truth mask, and the output of different methods: Baseline U-Net, MC, CRC, and MC-CP.}
\label{fig:qualitative_comparison}
\Description[Pixel- and instance-level metrics for comparative methods on the two evaluated datasets.]{Pixel- and instance-level metrics for comparative methods on the two evaluated datasets.}
\end{figure*}

\begin{table*}[tb!]
\centering
\caption{Evaluation metrics of \approach\ and other approaches for DeepMoon and YTU-WaterNet, reported at the pixel- and instance-level. The YTU-WaterNet--Instance-level metrics are omitted because only one hazardous area exists.}
\vspace{-4mm}
\label{tab:core-metrics}
\resizebox{\textwidth}{!}{%
\begin{tabular}{@{}ccccccccc@{}}
\toprule
\textbf{} & \textbf{} & \multicolumn{4}{c}{\textbf{Pixel-level Metrics}} & \multicolumn{3}{c}{\textbf{Instance-level Metrics}} \\ 
\cmidrule(lr){3-6} \cmidrule(lr){7-9}
\textbf{Dataset} &
  \textbf{Technique} &
  \textbf{Precision} &
  \textbf{Coverage} &
  \textbf{F1 (Dice)} &
  \textbf{IoU} &
  \textbf{Precision} &
  \textbf{Coverage} &
  \textbf{F1 Score} \\ \midrule
\multirow{4}{*}{DeepMoon} 
& Baseline & $44.59 \pm 21.68$ & $10.42 \pm 8.77$ &       $15.71 \pm 11.91$ & $8.98 \pm 7.18$ & $64.30 \pm  20.76$ & $60.49 \pm 16.95$ & $59.59 \pm 14.28$ \\
& MC & $48.26 \pm 28.19$ & $6.15 \pm 6.01$   & $10.38 \pm 9.56$  & $5.75 \pm 5.49$  & $67.69 \pm 20.38$ & $53.95 \pm 19.60$ & $56.29 \pm 14.83$ \\
& CRC & $26.63 \pm 8.08$  & $42.97 \pm 18.34$ & $30.38 \pm 10.76$ & $18.35 \pm 7.02$ & $43.53 \pm 16.14$ & $72.14 \pm 13.93$ & $52.43 \pm 11.68$ \\
& MC-CP & $25.72 \pm 4.92$  & $69.29 \pm 16.16$ & $25.36 \pm 7.34$  & $14.72 \pm 4.70$ & $35.67 \pm 6.35$  & $79.66 \pm 10.40$ & $45.54 \pm 8.60$  \\ 
\bottomrule
\multirow{4}{*}{YTU-WaterNet} 
& Baseline & $95.88 \pm 16.53$ & $94.78 \pm 15.91$ & $94.90 \pm 16.04$ & $92.93 \pm 17.54$ & \cellcolor{gray!40} &\cellcolor{gray!40} &\cellcolor{gray!40} \\ 
& MC & $95.86 \pm 16.60$ & $94.84 \pm 15.72$ & $94.94 \pm 16.00$ & $92.97 \pm 17.44$ & \cellcolor{gray!40} &\cellcolor{gray!40} &\cellcolor{gray!40} \\ 
& CRC & $86.04 \pm 19.91$ & $98.71 \pm 6.72$ & $90.21 \pm 15.94$ & $85.13 \pm 20.59$ & \cellcolor{gray!40} &\cellcolor{gray!40} &\cellcolor{gray!40} \\ 
& MC-CP & $72.88 \pm 27.22$ & $99.69 \pm 1.14$ & $80.77 \pm 22.39$ & $72.80 \pm 27.22$ & \cellcolor{gray!40} &\cellcolor{gray!40} &\cellcolor{gray!40} \\ 
\bottomrule
\end{tabular}%
}
    \vspace{-4mm}
\end{table*}

We evaluate \approach\ on two satellite-derived hazard segmentation benchmarks: the DeepMoon crater detection dataset~\cite{silburt2019lunar} and the YTU-WaterNet water segmentation dataset~\cite{erdem2021ensemble}. These datasets are representative of visually distinct but safety-critical terrain types: craters in planetary exploration and water bodies in terrestrial navigation. Both environments pose different perceptual challenges, i.e., subtle, low-contrast features in lunar imagery versus large, heterogeneous structures in aerial water segmentation, making them well-suited to assess \approach\ under varied conditions.  

We instantiate \approach\ with two conformal prediction algorithms:
Conformal Risk Control (CRC)~\cite{angelopoulos2022conformal}
and Monte-Carlo Conformal Prediction (MC-CP)~\cite{bethell2024robust}. 
This diversity in CP  instances demonstrates that our perception-to-policy pipeline is agnostic to the selected CP method and can accommodate different conformal segmentation approaches. 
For completeness, we also compared our approach against non-conformal baselines, i.e.,  a standard U-Net segmentation model and a U-Net with Monte Carlo dropout. 
Taken together, these baselines span deterministic, probabilistic, and conformal approaches, enabling a controlled assessment of how distribution-free calibration impacts both hazard perception and safe navigation. 

We use the following metrics to evaluate the approaches: \\
$\bullet$ precision: the proportion of hazardous pixels correctly identified, penalising false positives; \\
$\bullet$ coverage: the number of hazardous pixels correctly identified, penalising missed hazards; \\
$\bullet$ F1: the harmonic mean of precision and coverage; and\\ 
$\bullet$ IoU: the spatial overlap between predictions and ground truth. 

We report these metrics (average and standard deviation) over five independent runs for both pixel-level (image view) and instance-level (vector view).
DeepMoon has several crater instances within a single image; YTU-WaterNet has a single hazard instance (Figure~\ref{fig:qualitative_comparison}).

\begin{figure*}[tb]
\centering
    \begin{subfigure}{\textwidth}
    \includegraphics[width=\linewidth]{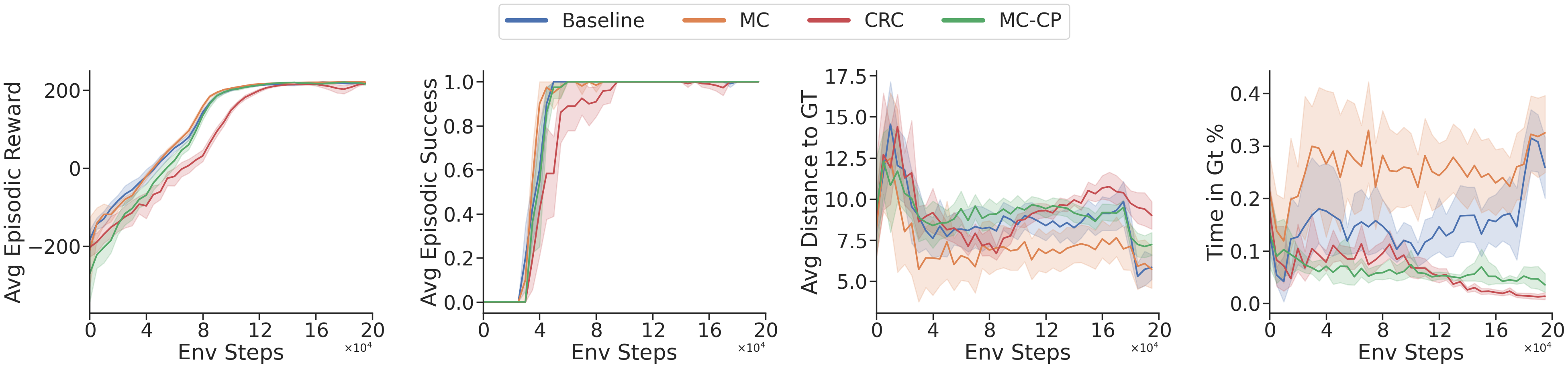}
    \vspace{-6mm}
    \caption{Environment 1 results}
    \label{fig:environment-7-results}
    \end{subfigure}
    
    \begin{subfigure}{\textwidth}
    \includegraphics[width=\linewidth]{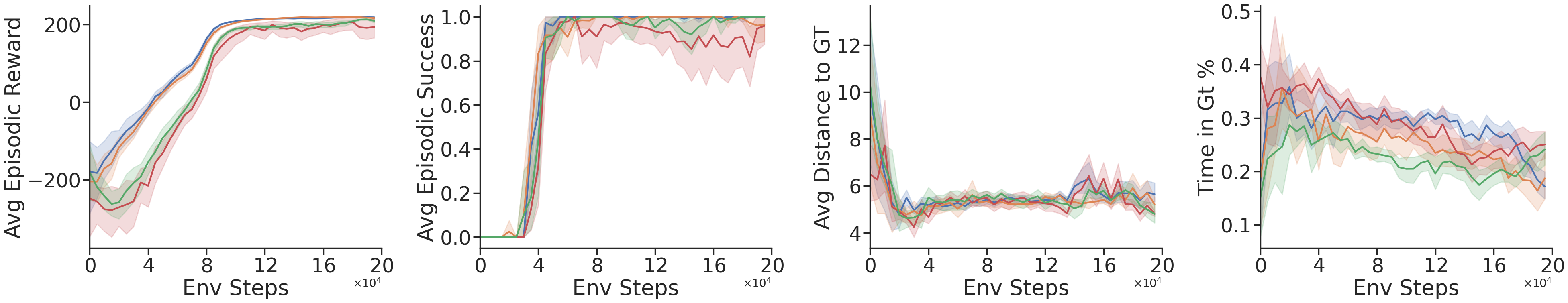}
    \vspace{-6mm}
    \caption{Environment 2 results}
    \label{fig:environment-3-results}
    \end{subfigure}

    \begin{subfigure}{\textwidth}
    \includegraphics[width=\linewidth]{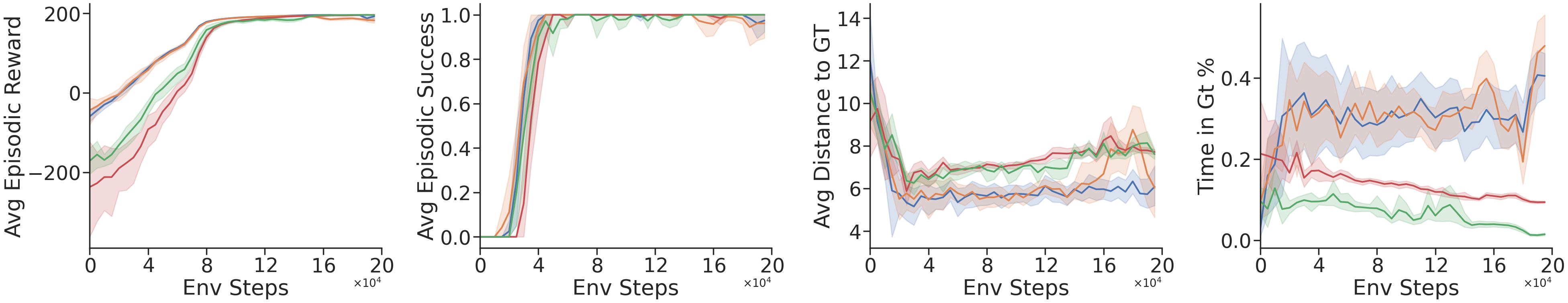}
    \vspace{-6mm}
    \caption{Environment 3 results}
    \label{fig:environment-12-results}
    \end{subfigure}

    \begin{subfigure}{\textwidth}
    \includegraphics[width=\linewidth]{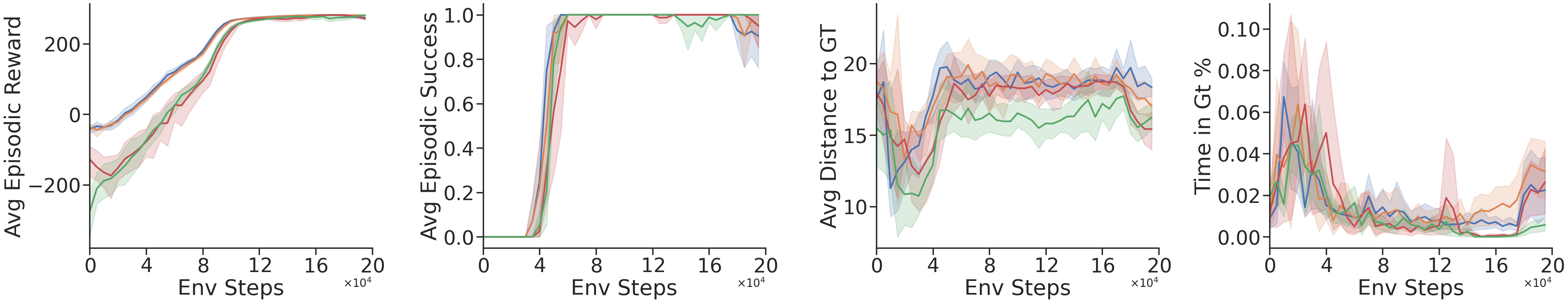}
    \vspace{-6mm}
    \caption{Environment 4 results}
    \label{fig:environment-10-results}
    \end{subfigure}
    \vspace{-6mm}
\caption{Performance of comparative methods on four DeepMoon environments (vector view). Curves show average episodic reward, success rate, distance to ground truth hazard, and time spent in ground-truth hazards for Baseline, MC, CRC, and MC-CP (shaded bands indicate variability).}
\label{fig:deepmoon-rl-results}
\Description[Fly 1 and Fly 2 look identical]{Fly 1 and fly 2 comparison shows identical length, wingspan, and overall bodily structure.}
    \vspace{-4mm}
\end{figure*}

\subsection{Core Results}
\label{sec:core-results}

Table~\ref{tab:core-metrics} summarises the core results from the conformalized hazard prediction stage across both datasets; examples are visualised in Figure~\ref{fig:qualitative_comparison}. On DeepMoon, the baseline U-Net achieves relatively high precision but extremely low coverage ($60.49 \pm 16.95$), often missing crater rims completely. 
Specifically, it correctly predicts approximately 10\% of the hazards at the pixel-level, and 60\% of the craters at instance-level (vector view). 
Monte Carlo dropout (MC) maintains similar behaviour, offering no improvement in coverage ($53.95 \pm 19.60$), but an improvement in precision. 
In contrast, \approach\ using either CRC or MC-CP substantially improves upon hazard predictions, achieving four to six times higher coverage. For example, using MC-CP, \approach\ correctly identifies approximately 69\% and 80\% of the craters using image and vector views, respectively, which are substantial improvements. 
Unavoidably, this performance gain comes at the cost of reduced precision, but the resulting predictions ensure that critical hazards are consistently captured at both pixel- and instance-level. 
Looking at the examples in Figure~\ref{fig:qualitative_comparison}, it is clear that the reduced precision of \approach\ arises due to a combination of predicted false positives and expanded hazard predictions (e.g., larger predicted craters than the ground truth). 
This outcome is expected as the CP-based \approach\ instances yield increased conservative estimates as the uncertainty becomes higher. 
However, in safety-critical environments, like those in our evaluation, where remote control from operators is challenging, if not impossible, the adoption of calibrated conservative behaviour enables the RL agent to operate within safety boundaries that satisfy the strict safety requirements imposed by the mission. 

Similarly, on the YTU-WaterNet dataset, the baseline U-Net, unlike on the DeepMoon dataset, achieves strong precision and coverage of $95.88 \pm 16.53$ and $94.78 \pm 15.91$, respectively. 
We observe similar trends in this dataset; \approach, with CRC and MC-CP, pushes hazard coverage close to perfect scores, evidencing its competency; as before, precision drops slightly due to false positives. 
Looking at the examples in Figure~\ref{fig:qualitative_comparison}, just like in DeepMoon, the baseline U-Net fails to accurately predict the hazardous regions, in this case, largely predicting an area of water as safe (land). 
\approach\ using CRC, instead, accurately covers the hazardous region close to the ground truth, while using MC-CP, it acts conservatively, adding a safety buffer around the hazardous region to ensure coverage.

The core results of the full \approach\ approach (including the RL path planner) are reported in Figure~\ref{fig:deepmoon-rl-results} for four DeepMoon environments over four metrics: average episodic reward, success rate, distance-to-goal, and time spent in ground-truth hazardous regions. 
Across all environments, both \approach\ variants (CRC, MC-CP) consistently outperform the non-conformal baselines.

\begin{figure*}[tb]
\centering
    \begin{subfigure}{\textwidth}
    \includegraphics[width=\linewidth]{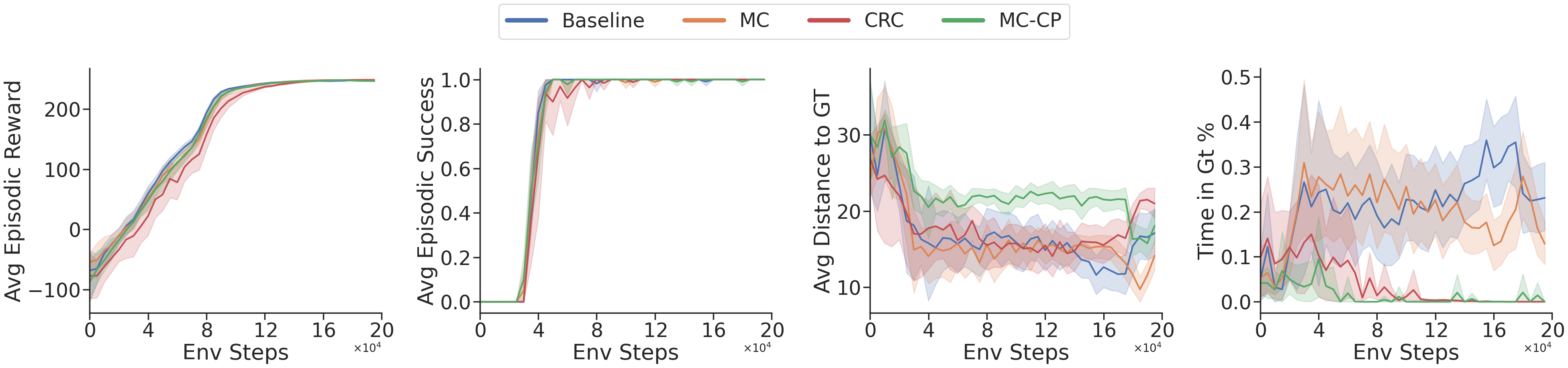}
    \vspace{-5mm}
    \caption{Environment 1 results}
    \label{fig:environment-12-results}
    \end{subfigure}
    
    \begin{subfigure}{\textwidth}
    \includegraphics[width=\linewidth]{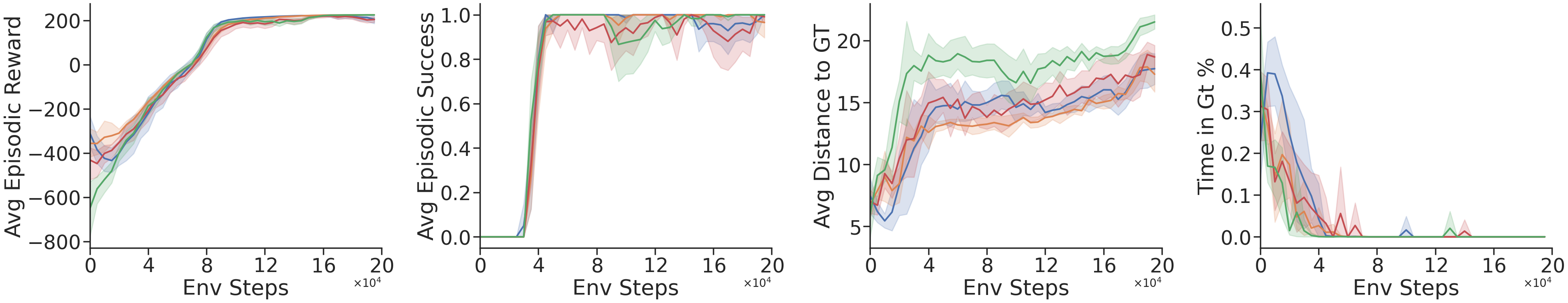}
    \vspace{-5mm}
    \caption{Environment 2 results}
    \label{fig:environment-01-results}
    \end{subfigure}
    
    \begin{subfigure}{\textwidth}
    \includegraphics[width=\linewidth]{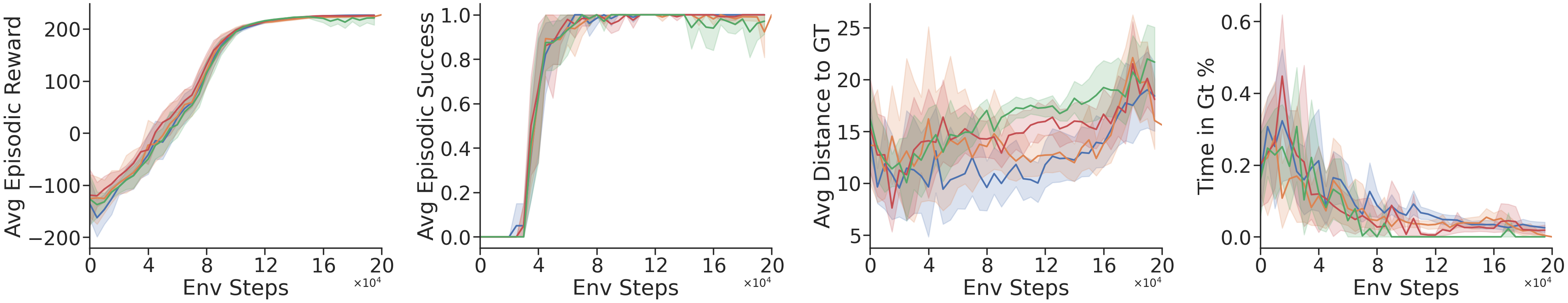}
    \vspace{-5mm}
    \caption{Environment 3 results}
    \label{fig:environment-09-results}
    \end{subfigure}

    \begin{subfigure}{\textwidth}
    \includegraphics[width=\linewidth]{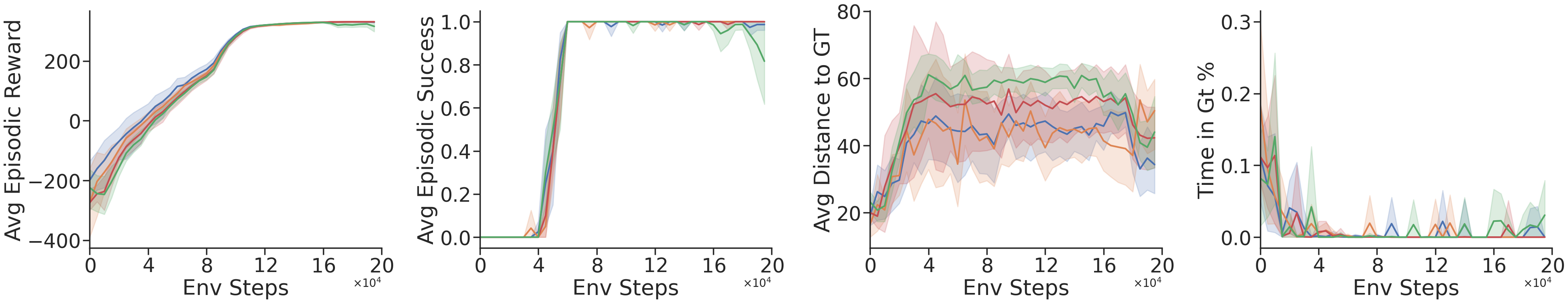}
    \vspace{-5mm}
    \caption{Environment 4 results}
    \label{fig:environment-05-results}
    \end{subfigure}
    \vspace{-6mm}
\caption{Performance of comparative methods on four YTU-WaterNet environments (vector view). Curves show average episodic reward, success rate, distance to ground truth hazard, and time spent in ground-truth hazards for Baseline, MC, CRC, and MC-CP (shaded bands indicate variability).}
\label{fig:waternet-rl-results}
\Description[Fly 1 and Fly 2 look identical]{Fly 1 and fly 2 comparison shows identical length, wingspan, and overall bodily structure.}
    \vspace{-4mm}
\end{figure*}

All methods eventually reach an average reward of $\approx 200$ and near-saturated success, indicating that introducing calibrated hazard costs does not hinder learnability. Despite an initial gap early in training, convergence time is comparable across methods, showing that conservative masks do not slow progression toward optimal returns. The average episodic success rate corroborates these insights as in all tested environments, after $\approx 40,000$ training steps, all four methods begin to reach an average episodic success rate of $1$.

The advantages of employing \approach\ are mostly evident in safety metrics. 
Specifically, the conformal variants maintain consistently larger clearance from ground-truth hazards across all four environments. After the initial learning phase, CRC and MC-CP achieve higher distance to the closest hazard with reduced variance, indicating more reliable conservative behaviour. For example, in Environment~3, the Baseline- and MC-based RL agents drift closer to crater rims as training progresses, averaging $\approx 6$-pixels from the nearest crater. 
In contrast, \approach's CRC- and MC-CP-based conservative behaviour average $\approx 8$-pixels from the nearest crater. 
This behaviour stems from the calibrated masks capturing crater rims more completely, shaping a cost field that is more accurate to the ground truth, and in MC-CP's case, creating an additional safety buffer around the hazards.

\begin{figure*}[tb]
    \centering
    \includegraphics[width=\linewidth]{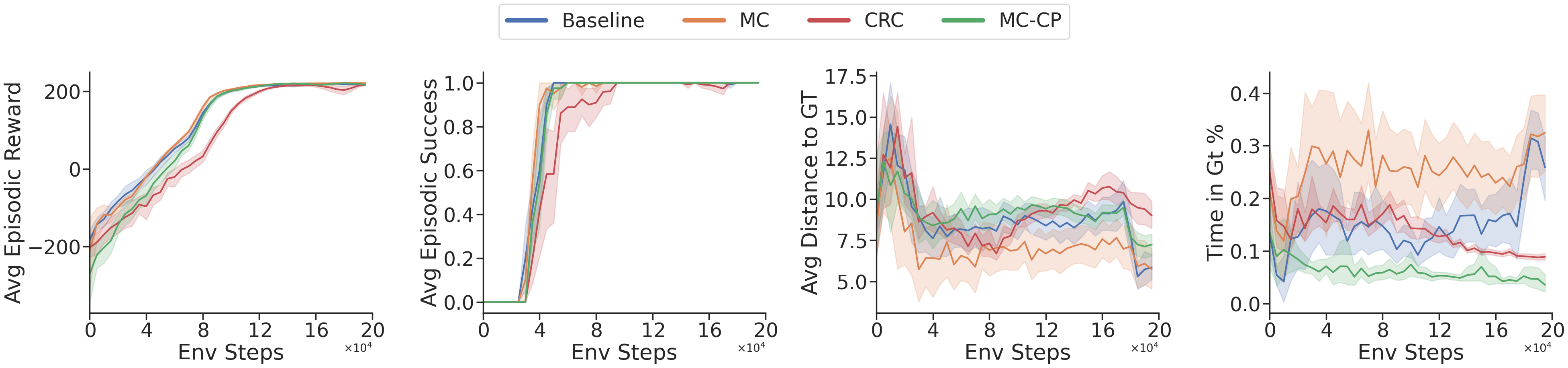}
    \vspace{-6mm}
    \caption{Performance of comparative methods on Environment 1 from the DeepMoon environments using the whole predictive hazard map as the state observation (image view). Curves show average episodic reward, success rate, distance to ground truth hazard, and time spent in ground-truth hazards for Baseline, MC, CRC, and MC-CP (shaded bands indicate variability).}
    \label{fig:deepmoon-image-results}
    \Description[A high-level overview of our proposed framework.]{A high-level overview of our proposed framework.
    }
    \vspace{-2mm}
\end{figure*}

\begin{figure*}[tb]
    \centering
    \includegraphics[width=0.95\linewidth]{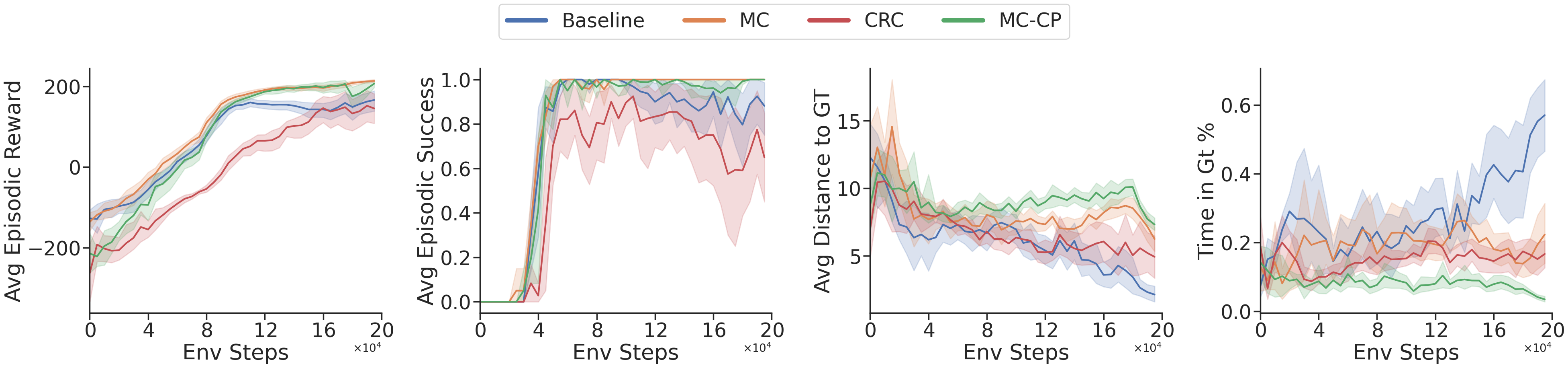}
    \vspace{-4mm}
    \caption{Performance of comparative methods on a state-wise noisy variant of Environment 1 from the DeepMoon environments (vector view). Curves show average episodic reward, success rate, distance to ground truth hazard, and time spent in ground-truth hazards for Baseline, MC, CRC, and MC-CP (shaded bands indicate variability).}
    \label{fig:deepmoon-noisy-results}
    \Description[A high-level overview of our proposed framework.]{A high-level overview of our proposed framework.
    }
    \vspace{-2mm}
\end{figure*}

A similar safety advantage is reflected in the fraction of time inside ground-truth hazardous regions (righmost column in Figure~\ref{fig:deepmoon-noisy-results}).
In Environment~3, for instance, the Baseline and MC trajectories spent on average  a third of their training time within the hazardous regions. 
\approach\ substantially reduces this to a fifth in the case of CRC due to more accurately predicting ground truth hazards in their predicted hazard map. MC-CP further boosts safety by increasing the coverage of the hazards, as seen in Table~\ref{tab:core-metrics}, and thus the time spent within the ground truth in this case was $\approx 0.075\%$ on average and often close to $\approx 0\%$. 
This result highlights the effectiveness of increasing coverage of the predicted hazards and the contribution of \approach. 
Even though, precision of the predicted hazards drops slightly for \approach\ instances  due to false positives, the effect on downstream agents is negligible in performance, and yielding significantly safety improvements across all environments.

Taken together with the perception results, these findings demonstrate that conformal segmentation (CRC or MC-CP) coupled with RL yields markedly safer behaviour, often approaching perfect safety, without sacrificing goal-reaching efficiency.

The same pattern holds on YTU-WaterNet (Figure~\ref{fig:waternet-rl-results}):  
\approach\ variants preserve reward and success while further reducing time in hazardous regions, confirming that our approach is robust across visually distinct hazard modalities. 
Even in scenarios such as Environment 2 in Figure~\ref{fig:waternet-rl-results}, where the baseline model can accurately enough ensure safety and reduce time spent in the hazardous region to near zero, \approach\ still provides extra safety by increasing its distance to the hazards beyond what the baseline can. 
This added safety is due to the extra safe spatial buffer around the hazards.

Using the full predicted hazard map as the observation (Figure~\ref{fig:deepmoon-image-results}), yields results consistent with the vector setting: all methods achieve similar reward and near-saturated success. The safety gap persists, with CRC and MC-CP maintaining larger clearance and lower, more stable time in hazardous regions, whereas Baseline and MC show larger variance and irregular behaviour especially at the latest mission stages. 
The richer spatial context incurs slightly slower early learning, but convergence remains comparable, confirming that \approach\ delivers the same safety gains regardless of observation type (image vs vector view). 

\subsection{Robustness to Distribution Shift}
\label{sec:Robustness to Distribution Shift}

To assess robustness under degraded sensing, we evaluate a state-wise noisy variant of DeepMoon Environment~1 (Figure~\ref{fig:deepmoon-noisy-results}). Noise is injected at the observation level by adding zero-mean Gaussian perturbations to the four distance-derived features with probability~$p$. Concretely, for the entries corresponding to goal distance and predicted–hazard distance (and their companion distance features), at each timestep we apply $o_i \leftarrow \mathrm{clip}\big(o_i + \varepsilon_i,\,[\ell_i, h_i]\big)$ with $\varepsilon_i \sim \mathcal{N}(0,\sigma^2)$, and we expand the bounds by $3\sigma$ so that the observation space remains truthful. In our setting, $\sigma = 1.0$ and $p = 1.0$, so all four distance entries are perturbed at every step.

Despite noisy distances, \approach\ with either CRC or MC-CP retains high reward and success comparable to the best baseline while showing markedly safer behaviour. For the average distance to the nearest hazard, MC-CP maintains the largest average clearance with narrow variability bands showing the additional buffer given by MC-CP makes our approach robust to noisy data, an essential feature in deployment settings; CRC tracks closely. 
In contrast, Baseline and MC drift closer to crater rims as training proceeds and exhibit visibly higher variance. The difference is even clearer in the time spent in the hazards: the baseline agents spend a higher fraction of time within ground-truth hazards and show late-episode spikes, whereas MC-CP remains exhibits stable behaviour, with CRC consistently below both baselines. 
These trends clearly indicate that distribution-free calibrated masks create a risk-aware cost field that continues to steer RL policies away from hazards even when the state features that approximate distances are uncertain or corrupted, yielding safety gains without sacrificing performance.


\section{Conclusion}
\label{sec:Conclusion}

We introduced \approach, a perception-to-policy learning approach that brings distribution-free, finite-sample safety guarantees into navigation by calibrating hazard segmentation and exposing a risk-aware cost field to RL agents. 
Across both lunar and terrestrial benchmarks, our conformal variants (CRC and MC-CP) consistently reduce time in hazardous regions and increase clearance from unsafe terrain while maintaining task performance and convergence comparable to non-conformal baselines.
\approach\ is modular, agnostic to the conformal calibrator choice, and robust to distributional shift. 
These results indicate that coupling statistically calibrated perception with RL yields substantially safer behaviour without compromising efficiency, offering a practical path toward reliable autonomous navigation in safety-critical settings.
Future work will explore extending \approach\ to multi-agent and partially-observable RL agents, also investigating online calibration and adaptive risk estimation to further enhance scalability and operational reliability.


\begin{acks}
To Robert, for the bagels and explaining CMYK and color spaces.
\end{acks}


\printbibliography

\end{document}